\documentclass[11pt,letterpaper]{article}
\usepackage{cogsys}
\usepackage[T1]{fontenc}
\usepackage{times}
\usepackage[pdftex]{graphicx} 
\usepackage{subcaption} 

\usepackage{natbib}
\cogsysheading{X}{20XX}{1-6}{X/20XX}{X/20XX}

\ShortHeadings{Incremental Concept Formation without Catastrophic Forgetting}
              {N.\ Barari, X.\ Lian, and C.\ J.\ MacLellan}

\begin{document} 

\title{Incremental Concept Formation over Visual Images Without Catastrophic Forgetting}
 
\author{Nicki Barari$^*$}{nicki.barari@drexel.edu}
\address{Drexel University, 
         Philadelphia, PA 19104 USA}
\author{Xin Lian$^*$}{xinlian@gatech.edu}
\author{Christopher MacLellan}{cmaclell@gatech.edu}
\address{Teachable AI Lab, Georgia Institute of Technology, Atlanta, GA 30332 USA}

\vskip 0.2in

\def\thefootnote{*}\footnotetext{These authors contributed equally to this work.}\def\thefootnote{\arabic{footnote}}
 
\begin{abstract}Deep neural networks have excelled in machine learning, particularly in vision tasks, however, they often suffer from catastrophic forgetting when learning new tasks sequentially. In this work, we introduce \textit{Cobweb/4V}, 
an alternative to traditional neural network approaches.
Cobweb/4V is a novel visual classification method that builds on Cobweb, a human-like learning system that is inspired by the way humans incrementally learn new concepts over time.
In this research, we conduct a comprehensive evaluation, showcasing Cobweb/4V’s proficiency in learning visual concepts, requiring less data to achieve effective learning outcomes compared to traditional methods, maintaining stable performance over time, and achieving commendable asymptotic behavior, without catastrophic forgetting effects. These characteristics align with learning strategies in human cognition, positioning Cobweb/4V as a promising alternative to neural network approaches. 
\end{abstract}

\section{Introduction} 
 
Computer vision has made substantial progress in recent years due to breakthroughs in deep neural networks. Neural networks can exceed human capabilities in certain tasks such as object detection and classification \citep{he2016deep}. However, such networks cannot handle continual learning of new tasks without forgetting previously learned data. Catastrophic forgetting is a fundamental challenge for artificial neural networks \citep{mccloskey1989catastrophic}. This phenomenon happens when the network is trained on multiple tasks sequentially, and to meet the objective of the new task, it changes the weights learned to perform the previous tasks. To prevent it, the model needs to be trained on both new data and previous data simultaneously, which consumes more time and energy, underscoring the trade-off between efficiency and performance in neural networks. 

As humans, we excel at learning knowledge and skills across different domains, adapting them through new experiences, and transferring them throughout our lifespan \citep{barnett2002and, calvert2004handbook}. While we gradually forget some of our previously learned knowledge over time, it rarely happens that our existing knowledge catastrophically fades away when learning something new. To achieve more efficiency, while maintaining high performance, neural networks need new human-like capabilities for incrementally learning new concepts over time \citep{french1999catastrophic}. Over the past two decades, extensive research has been dedicated to studying catastrophic forgetting, aiming to mitigate its impact on neural networks. These approaches can be categorized into four main groups, 1) regularization-based, 2) replay-based, 3) optimization-based, and 4) architecture-based methods \citep{wang2023comprehensive}. While contributing significantly to mitigating this phenomenon, these methods come with limitations including reliance on memory-intensive mechanisms, difficulties in preserving old parameters when dealing with a growing number of tasks, and sensitivity to task-specific parameters. 

In this work we present \textit{Cobweb/4V}, a novel visual classification approach that builds upon \textit{Cobweb} \citep{fisher2014concept,gennari1989models}, a concept formation technique that draws inspiration from the psychological study of how humans learn concepts. Cobweb supports all of the human-like learning capabilities outlined in \citep{langley2022computational}, such as learning rapidly in an incremental, piecemeal way, while explaining basic-level and typicality effects from psychology \citep{fisher1988computational}. This work aims to combine computer vision principles with this prior approach and explore the idea of incorporating new visual information incrementally without erasing previously learned data. Our results demonstrate that Cobweb/4V does not exhibit catastrophic forgetting, only limited interference effects when compared to neural networks. We find that Cobweb/4V is competitive with neural network approaches while having minimal forgetting effects. It is also more data efficient and achieves asymptotic performance with fewer examples. Overall, Cobweb/4V's capacity to preserve previously acquired knowledge while incrementally incorporating new experiences makes it a promising alternative to neural networks. Moreover, it provides insights into learning mechanisms that closely resemble human processing, particularly when applied to vision tasks.

\section{Related Works}
\subsection{Continual, Lifelong Learning}

\textit{Continual learning} refers to the ability of a learning model to adapt to new tasks without forgetting previously learned knowledge. It is an important area of research in machine learning as it helps machine learning models to become more human-like by being able to accumulate knowledge over their lifetime, improves data efficiency, and enhances generalization capabilities by applying past knowledge to new, unseen tasks, and more importantly, addresses the challenge of catastrophic forgetting, where models tend to lose previously learned information when exposed to new tasks.   
Catastrophic forgetting happens due to a trade-off between learning plasticity and memory stability. Earlier efforts to explore continual learning \citep{mccloskey1989catastrophic} laid the foundation for addressing catastrophic interference in connectionist networks, highlighting the challenges of sequential learning and the need for adaptive algorithms in machine learning. Since then, extensive research has been dedicated to addressing this fundamental challenge, which can be categorized into four main categories, the replay-, optimization-, regularization-, and architecture-based methods \citep{wang2023comprehensive}.

\textbf{\textit{Replay-based}} methods tackle catastrophic forgetting by revisiting data from past experiences during the learning of new tasks. Early approaches, such as rehearsal in backpropagation networks \citep{robins1993catastrophic}, underscored the advantage of revisiting previously learned patterns. Following that, the development of the pseudo-rehearsal technique \citep{robins1995catastrophic} was prompted by scenarios in which old training data was not accessible, which led to the adoption of a memory network for generating pseudo-examples during the retraining. The Generative Adversarial Networks (GANs) \citep{goodfellow2014generative} led to a new approach called deep generative replay \citep{shin2017continual}, which employs a dual-model architecture with a generator based on GANs for generating ``fake'' inputs from the previous training data distribution and a solver for image classification. 
However, replay methods often involve revisiting past examples, which increases computational costs and training time. In addition, capturing the diversity of past experiences can be challenging, with generative replay struggling to reproduce a wide range of learning scenarios, potentially leading to incomplete learning and reduced performance.

\textbf{\textit{Optimization-based}} methods offer a new paradigm to address the limitations associated with replay, by adjusting the optimization program. Gradient Episodic Memory (GEM) \citep{lopez2017gradient} is an example that integrates samples from earlier tasks into the loss function, using an episodic memory to store these samples, so it effectively balances adapting to new information and preserving knowledge from prior tasks without requiring access to previous models. However, GEM relies on task boundaries during training and experiences performance degradation with reduced memory size. As more memories accumulate, the optimization problem becomes more constrained, negatively impacting knowledge transfer to later tasks.

\textbf{\textit{Regularization-based}} methods enhance a model's adaptation to new tasks while maintaining previous knowledge by adding regularization penalties and constraints to the training loss function to penalize significant changes to crucial parameters. The weight regularization method preserves critical parameters for prior tasks to reduce catastrophic forgetting, such as Elastic Weight Consolidation (EWC) \citep{kirkpatrick2017overcoming}, which estimates the importance of each weight regarding previous tasks using the Fisher information matrix. The function regularization method targets a function's intermediate or final output, such as Learning without Forgetting (LwF) \citep{li2017learning}, which reduces forgetting by introducing a regularization term and computes the distillation loss \citep{hinton2015distilling}. 

\textbf{\textit{Architecture-based}} methods involve designing model architectures that inherently support continual learning. One example is the Progressive Neural Networks (PNNs) \citep{rusu2016progressive} which employs modular networks, and adds new columns or neural networks for each subsequent task while maintaining existing ones. This ensures the forward transfer of learned features from one network to another, preventing interference with learning without forgetting. However, as tasks increase, a PNN's complexity grows, leading to higher computational demands and inefficiencies in memory and computational resources \citep{parisi2019continual}. Additionally, PNNs lack positive backward transfer, meaning knowledge from newer tasks cannot enhance performance on earlier tasks \citep{schwarz2018progress}.

While continual learning is a pivotal step towards achieving human-like learning, it encounters various challenges, including the widespread issue of catastrophic forgetting that extends across different approaches such as regularization and optimization-based methods \citep{kirkpatrick2017overcoming, li2017learning}. Additionally, issues related to scalability and computational cost are evident in replay- and architecture-based methods \citep{robins1995catastrophic, rusu2016progressive}. In our evaluation of Cobweb/4V, we draw insights from a recent study \citep{van2019three} that extensively explored continual learning approaches and bring attention to their limitations, particularly in scenarios involving the learning of new classes, notably with regularization-based methods. To align our evaluation with best practices, we choose to benchmark Cobweb/4V against a replay-based approach, highlighting the widely recognized effectiveness of replay in shaping successful continual learning strategies.

\subsection{Cobweb} 
\begin{figure}[t!]
\centering
\includegraphics[width=0.6\textwidth]{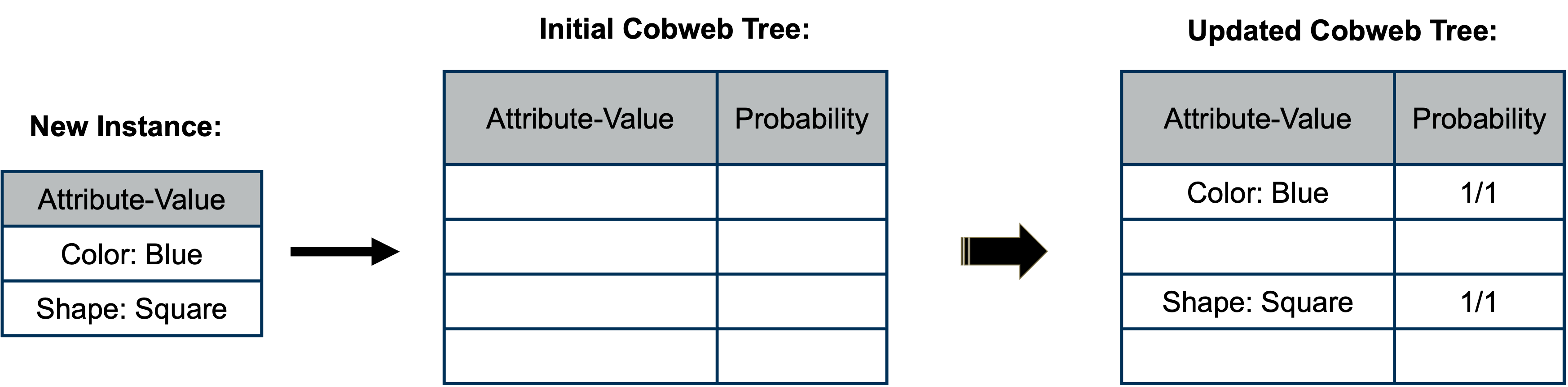}
\caption{A new instance is incorporated into an empty Cobweb tree. Cobweb adds it to the root and updates the root's count table to reflect the instance's attribute-values.}
\label{fig:cobweb_instance}
\end{figure}

This approach presents an alternative to neural networks, supporting incremental and unsupervised learning over a stream of examples \citep{fisher1987knowledge,fisher2014concept}. Given consecutive instances, {\it Cobweb} learns a concept hierarchy in a fashion inspired by how humans form concepts. Each instance is represented by a set of discrete attribute-value pairs,
e.g. \texttt{\{color: blue; shape: square\}}, and each concept node in a Cobweb tree is represented by a table of attribute-values and probabilities reflecting the adopted instances (see Figure \ref{fig:cobweb_instance}). 

\begin{figure}[t!]
  \centering
  \begin{subfigure}[b]{0.6\textwidth}
    \includegraphics[width=\textwidth]{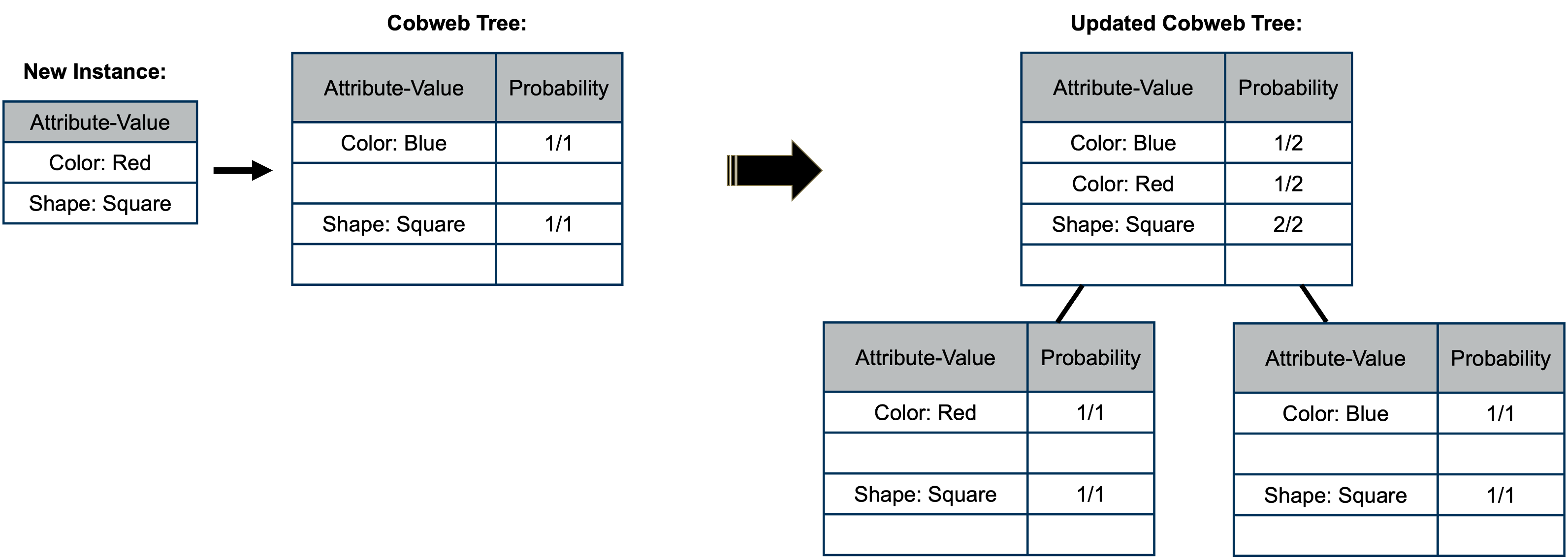}
    \caption{Cobweb categorizes a new instance down its hierarchy and updates the attribute value counts.}
    \label{fig:cobweb_learning}
  \end{subfigure}
  \hfill
  \begin{subfigure}[b]{0.35\textwidth}
    \includegraphics[width=\textwidth]{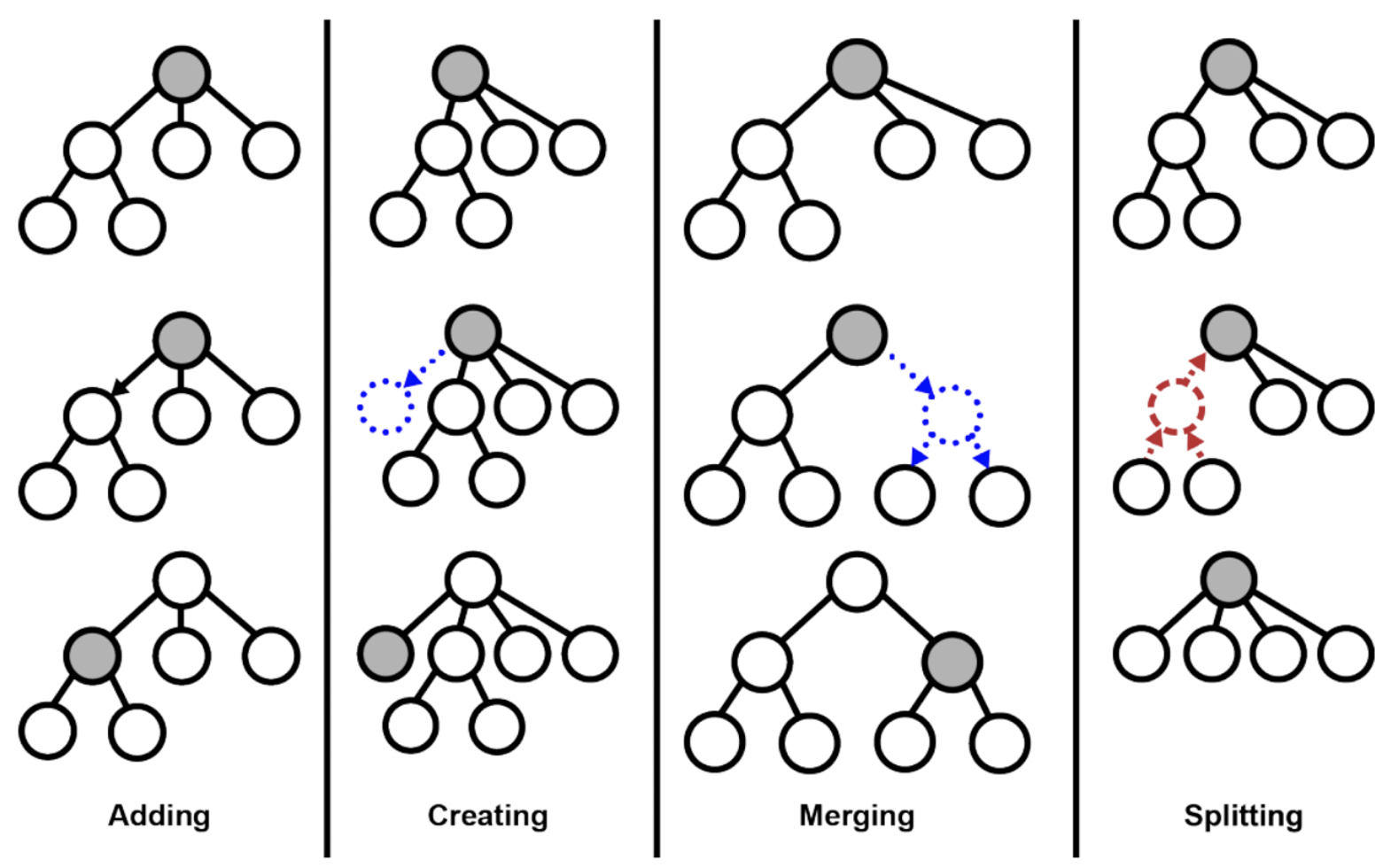}
    \caption{Four update operations that Cobweb considers at each node during learning \citep{maclellan2016trestle}.}
    \label{fig:cobweb_operations}
  \end{subfigure}
  \caption{Cobweb's Learning Process}
  \label{fig:main}
\end{figure}

During learning, Cobweb categorizes new instances down its tree recursively and updates the count tables of concepts along the categorization path to reflect the attribute values of the instances (Figure \ref{fig:cobweb_learning}). At each branch point, Cobweb considers four restructuring operations: \textit{adding} (adding the instances to the best child concept and updating the attribute value counts), \textit{merging} (combining the two ``best'' children and reconsidering the available operations), \textit{splitting} (removing the ``best'' child and elevating its children to the current level), and \textit{creating} (creating a new child node which adopts the new learned instance only) (see Figure \ref{fig:cobweb_operations} from \cite{maclellan2016trestle} ). 

During prediction, Cobweb follows a similar approach to learning, but it does not modify the counts of concepts along the path. Given an instance (potentially missing some of its attribute values), Cobweb sorts it downward through its hierarchy: Starting at the root node, at every branch (a concept node with its children), Cobweb recursively considers whether to direct the instance to one of the children or whether to stop at the current node. When this process terminates, it uses the final concept's count table to predict the values of unobserved attributes.

During both prediction and learning, Cobweb uses a measure called \textit{category utility} \citep{corter1992explaining} to evaluate its choices, and proceeds with the operation that maximizes category utility. Category utility evaluates the increase in the predictive power of a child node compared to its parent, and is similar to the information gain metric used in decision tree induction \citep{quinlan1986induction}; however, instead of supporting prediction of a single attribute, it has the capacity to support prediction of all attributes. The function is defined as:

\begin{equation}
    \frac{\sum_{k=1}^{n}P(C_k)\left[\sum_{i}\sum_{j}P(A_i=V_{ij}|C_k)^2-P(A_i=V_{ij})^2\right]}{n}
\end{equation}

Where $n$ is the number of concepts, $P(C_k)$ is the overall probability of the $k$th child concept, $P(A_i=V_{ij}|C_k)$ is the probability of attribute $A_i$ having value $V_{ij}$ given concept $C_k$, and $P(A_i=V_{ij})$ is the probability of attribute $A_i$ having value $V_{ij}$ in the parent concept. In general terms, $\sum_{i}\sum_{j}P(A_i=V_{ij}|C_k)^2$ represents the expected number of attributes correctly predicted for a given child $C_k$, while $\sum_{i}\sum_{j}P(A_i=V_{ij})^2$ corresponds to the expected number of correct guesses made from the parent. This equation represents the average increase in the number of correctly guessed attribute-values within the children compared to the parent, weighted by the probability of each child. To compare cases with different numbers of children, this score is  normalized by dividing by $n$, the number of children. 

While the original Cobweb approach only supports nominal attributes, \textit{Cobweb/3} \citep{mckusick1990cobweb} extends this support to continuous attributes by using normal distributions to describe their probability densities. Each concept stores the mean and standard deviation for each continuous attribute rather than the counts that are maintained for nominal attributes (see Figure \ref{fig:tensor}).

\section{Cobweb/4V} 
This paper introduces \textit{Cobweb/4V}, a new Cobweb variant designed for visual learning tasks. It extends a prior implementation by \cite{maclellan2016trestle} with a new prediction approach that uses multiple concepts throughout the categorization tree to yield improved predictive performance. It also introduces a new tensor-based concept representation that enables efficient processing of images.

\subsection{Learning with Mutual Information}
All prior Cobweb studies employed the \textit{probability-theoretic} category utility \citep{corter1992explaining} in both predicting and learning processes, which calculates the expected increase in the ability to predict attribute values correctly given the knowledge of an instance's membership in a particular concept. An analysis by \cite{fisher1996iterative} suggests this formulation is an unsupervised extension of the {\it Gini Index} measure commonly used to construct decision trees. In Cobweb/4V, however, we use the \textit{information-theoretic} category utility \citep{corter1992explaining}. This involves establishing a connection between feature predictability and informativeness. Using this new measure yields the following updated category utility function:

\begin{equation}
    \frac{\sum_{k=1}^{n}P(C_k)\left[H(A=V) - H(A=V|C_k)\right]}{n}
\end{equation}


where $H(A=V) = \sum_i \sum_j [-P(A_i=V_{ij}) \log(P(A_i=V_{ij})]$ is the entropy over all attribute values in the parent and $H(A=V|C_k)$ is the entropy over all attribute values in the given child $k$. This measure provides an unsupervised extension of the {\it information gain} measure commonly used in decision trees. It is analogous to mutual information, evaluating the increase in the information known about the attribute values gained by knowledge of the children over the parent. As the Gini index is a first-order approximation of information gain, this new variation of category utility supports higher precision than the probability-theoretic variation. Also, by formulating category utility in terms of entropy, we can easily use different distributions to represent attributes, as many commonly used distributions have closed-form formulas for computing their entropy. 

\subsection{Predicting with a Combination of Concepts}
During prediction, Cobweb first finds a final concept at the subordinate level in a greedy way, then uses the count information associated with this node to predict the unobserved attribute values of the categorized instance.
This approach has been widely adopted in prior studies because it seems to yield good empirical performance \citep{maclellan2016trestle, maclellan2022convolutional, Maclellan2022Efficient}. However, other studies have explored alternative concept levels for predictions, such as using  \textit{basic-level} concepts \citep{fisher1990structure, corter1992explaining}.

Our revised approach introduces a new prediction strategy that combines predictions from multiple concepts within the tree. Given an instance $x$ with some unobserved feature(s) and a maximum number of the concepts to use for prediction ($N_{max}$), Cobweb begins categorization at the root. However, instead of sorting $x$ down a single, greedy path, Cobweb/4V instead uses \textit{best-first} search. At each point in the search, it expands the node $c^*$ on the search frontier that has the greatest score $s(c) = P(c|x)P(x|c) \label{eqn:collocation}$. This heuristic, which is sometimes referred to as \textit{collocation} \citep{jones1983identifying},\footnote{Prior work shows it is related to probabilistic category utility; the expected correct guesses of a concept can be reinterpreted as a form of weighted collocation \citep{corter1992explaining}.} is the product of cue and category validity.
Once a $c^*$  is identified, the system adds it to an expanded node list. Next, the system evaluates the collocation of the children of $c^*$ and adds these to the search frontier to be considered for expansion.
This process repeats until Cobweb/4V has expanded $N_{max}$ nodes.
If $\mathcal{C}^*$ denotes all the nodes expanded during categorization, Cobweb/4V predicts the attribute $X_i$ will have value $x_i$ with probability:
\begin{equation}
    P(X_i=x_i|\mathcal{C}^*) = \sum_{c\in\mathcal{C^*}}P(x_i|c)\frac{\exp\{-s(c)\}}{\sum_{c\in\mathcal{C^*}}\exp\{-s(c)\}} \label{eqn:predict-value}
\end{equation}

\noindent This represents the combination of predictions from all expanded nodes, weighted by their collocation (the softmax ensures the sum of the weights to one). Although Cobweb/4V only expands $N_{max}$ nodes, rather than all the nodes, this approach is effectively a form of Bayesian model averaging \citep{hinne2020conceptual}. 

\begin{figure}[t!]
\centering
\includegraphics[width=0.7\textwidth]{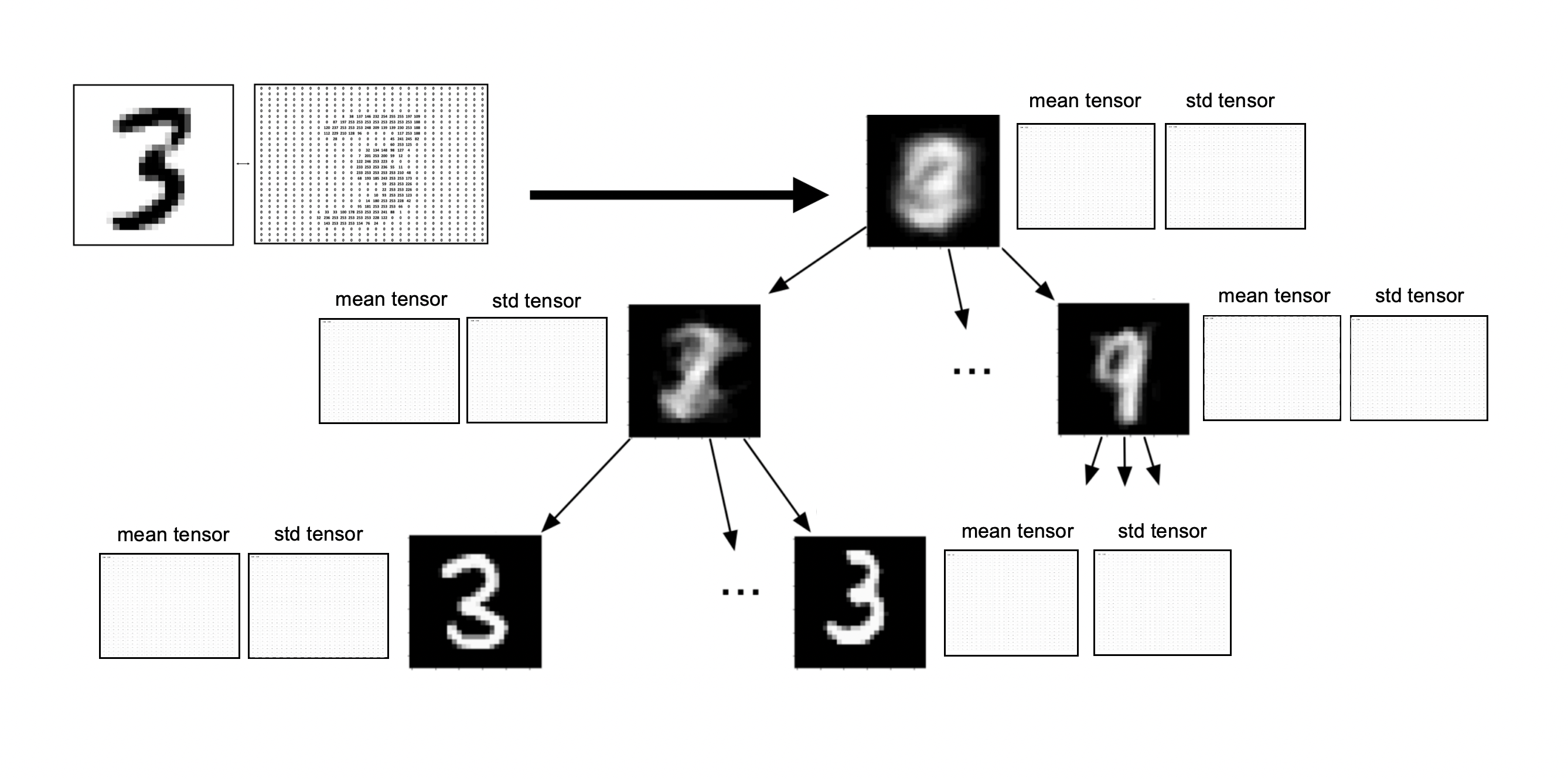}
\caption{The single image on the left shows the tensor-based representation for instances. The concepts store the means and standard deviations for the pixel attributes in internal tensor representations as well.}
\label{fig:tensor}
\end{figure}

\subsection{A New Tensor Representation}
In Cobweb/4V, each instance is represented by a tensor of pixels and a class label, as opposed to a list of attribute-value pairs, as used in previous Cobweb approaches. Given our focus on vision tasks, this adaptation better enables Cobweb/4V to support vision learning tasks, where each input takes the form of an $n$-channel 2D image, which is represented as a matrix of pixel intensity values.\footnote{We use grayscale images in our experiments, so the inputs are just a single channel (i.e., $n=1$), but the approach supports multi-channel input as well.} 
Each instance is also paired with a nominal digit label, which Cobweb/4V natively supports as a categorical, rather than requiring a specialized encoding, such as one-hot. Similar to Cobweb/3, each node stores the means and standard deviations of the pixel features for all the images that have been categorized under it. However, it uses a tensor representation to store this statistics, which enables more efficient inference and updating. Also, like Cobweb, it stores digit labels in a probability table that keeps track of how frequently each label appears across instances assigned to a given concept (Figure \ref{fig:tensor}).
The values for each attribute are assumed to be conditionally independent given the concept. Consequently, the uncertainty of a node $c$ is the sum of entropy values for the attributes under $c$. 
The use of tensor representations for instances and concepts (our implementation uses PyTorch) significantly enhances speed, yielding substantially faster processing than prior versions of Cobweb that use attribute-value lists. 

\section{Experiments}

We evaluate Cobweb/4V and two neural network baselines across distinct experimental conditions with the MNIST dataset (60,000 images in the training set, 10,000 images in the test set), prevalently used in the continual learning studies \citep{van2022three,lecun1998mnist}.\footnote{The codes for the experiments are available at \url{https://github.com/Teachable-AI-Lab/cobweb-vision}}

\subsection{Preliminary Experiment}
\begin{figure}[t!]
\centering
\includegraphics[width=0.8\textwidth]{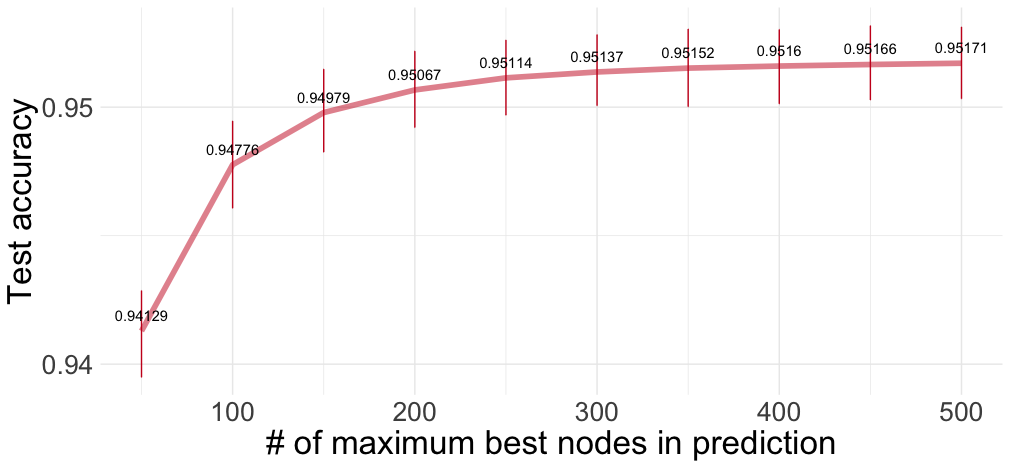}
\caption{The maximum best nodes considered in predictions combination vs. the averaged test accuracy with 95\% confidence intervals (evaluated on the entire MNIST test set) of Cobweb/4V after training on the entire MNIST training set.}
\label{fig:cobweb_asymptotic}
\end{figure}

The only hyperparameter for Cobweb/4V is the maximum number of concept nodes ($N_{max}$) to expand during prediction.
Given that our prediction strategy has never been tested before, we were unsure how performance would change as a result of increasing $N_{max}$.
We hypothesized that expanding more nodes would result in better predictions at increased computational cost, but there was a possibility that performance might be U-shaped, peaking at a particular value and then getting worse as more (generally less relevant nodes) are expanded.
To test our hypothesis and determine a suitable $N_{max}$ value for our later experiments, we trained Cobweb/4V using the entire MNIST training set and tested it with the complete test set, varying the $N_{max}$ value (10 times with different random seeds for each run). Our results, shown in Figure~\ref{fig:cobweb_asymptotic}, provide evidence to support our hypothesis that expanding more nodes increases predictive performance. Based on this test, we set $N_{max}$ to $300$ for the remaining experiments, as it has commendable performance (an accuracy of 0.951) and reasonable computation cost. 

\subsection{Baseline Neural Network Architectures}

We conducted a comprehensive performance comparison between Cobweb/4V and two baseline neural network architectures. The first one employs fully connected layers (\texttt{fc}), a versatile choice for visual classification with demonstrated effectiveness across multiple tasks \citep{rumelhart1986learning, lecun1998gradient, krizhevsky2009learning}. The second baseline incorporates additional convolutional neural network (CNN) layers (\texttt{fc-cnn}), which have proven effective for computer vision tasks \citep{lecun1998gradient, gu2018recent}.

To investigate catastrophic forgetting, we set the number of hidden layers in the \texttt{fc} network to 1 to be directly comparable to Cobweb/4V, which does not use a multi-layer architecture. The \texttt{fc-cnn} network is more representative of a typical CNN for MNIST, with ten $5\times5$ convolutional filters, a $2\times2$ max-pooling layer, twenty $5\times5$ convolutional filters, another $2\times2$ max-pool layer, and one fully connected layer.
To tune the hyperparameters, a grid search was employed with the \texttt{fc} network to identify optimal values.
Both \texttt{fc} and \texttt{fc-cnn} networks use ReLU activation functions and are trained with 5 epochs, 0.9 SGD momentum, a hidden-layer size of 128, a batch size of 64, and a learning rate of 0.00365.

\subsection{Methods and Hypotheses}

In the experiments, we evaluate \texttt{cobweb4v} and \texttt{fc(-cnn)} with the MNIST data under different tasks, each involving a set of sequential training splits with different compositions based on the different goals of the experiments.

\subsubsection{Experiment 1 (Learning with Modest Data)}

With this experiment, we aim to evaluate how Cobweb/4V's data efficiency in an incremental learning setting compares to the neural network baselines. To do this, we randomly shuffle the MNIST training set, divide it into splits of 10 examples each, and then present each split sequentially to each approach for training. We chose a split size of 10 because larger sizes made it difficult to see the rate of learning of each approach and smaller sizes introduced challenges for the neural network baselines.\footnote{For this experiment, each NN used a batch size of 5, so it could sample 2 batches from each split per epoch.} After training on each split, we evaluate each approach's accuracy on the entire MNIST test set.
To ensure robustness and account for randomness in the shuffling process, we ran the experiment 10 times using 10 different random seeds for shuffling on each occasion.  

We expect each approach to improve its performance given more training examples. Notably, we expect that the neural network baselines will require more data to achieve comparable performance to \texttt{cobweb4v}. This is attributed to the substantial number of gradient updates needed for neural networks to converge, whereas \texttt{cobweb4v} operates differently. Given the extensive MNIST training dataset and the success of CNNs on this dataset in prior work \citep{lecun1998gradient, krizhevsky2012imagenet}, we anticipate that \texttt{fc-cnn} may surpass \texttt{cobweb4v} at some point in the training process, ultimately achieving greater performance.

\subsubsection{Experiment 2 (Learning without Forgetting)}
This experiment aims to evaluate the extent to which each approach exhibits catastrophic forgetting. We begin by selecting one digit (ranging from 0 to 9) as the chosen digit label. We then divide the training set into 10 training splits. The first training split is constructed to include all available training data for the chosen digit, along with 600 images from each of the non-chosen digits (the remaining 9 digits). This ensures that each approach initially learns knowledge for all available digits. Subsequently, the remaining training data from all non-chosen digits is randomly divided across the remaining 9 splits. Similar to Experiment 1, we sequentially train each approach on each split, evaluating the approach exclusively with test set items labeled with the chosen digit after each split. To account for variability, we run each experiment 10 times using different random seeds for shuffling and splitting. Furthermore, we repeat the experiment 10 times, each time selecting a different chosen digit (resulting in $10 \times 100$ experimental runs for each approach).

Given that neural nets are known to exhibit forgetting, we also tested neural network variations that utilize {\it replay} to mitigate this effect.
Each replay approach (labeled \texttt{fc(-cnn)-replay}), maintains a replay buffer with 1,000 examples.
Given each new split, the network is trained with all the data in the replay buffer {\it and} all the data in the incoming split.
After training, we randomly sample 1,000 examples from the buffer and split data. This sample is then carried forward in the buffer for the next training iteration. Based on prior work by \cite{van2019three}, we chose to use replay because their analysis suggests it outperforms alternatives, such as regularization.\footnote{We also tested EWC, but found its performance comparable to the standard NN approaches, so we do not report it here.} Additionally, we employ sampling-based replay because other work suggests that rehearsing tasks on a fading schedule is particularly effective for mitigating forgetting \citep{gandhi2024catastrophic}. Our sampling approach emulates faded practice, as all digits (even the chosen one that only appears in the first split) have some fading probability of being rehearsed at all subsequent splits. 

We anticipate that \texttt{fc(-cnn)} will exhibit catastrophic forgetting of the chosen digit after the initial training split. As these approaches encounter subsequent training splits that exclude the chosen digit, their proficiency in classifying it is expected to rapidly decline. This phenomenon in neural networks is attributed to weight adjustments accommodating new task objectives, leading to the replacement of prior task knowledge \citep{kirkpatrick2017overcoming}. We expect \texttt{fc(-cnn)-replay} will be more robust to forgetting, but once the chosen digit data falls out of the finite buffer, it will also exhibit catastrophic forgetting. In contrast, \texttt{cobweb4v} is expected to maintain consistent performance on the chosen digit, even when being repeatedly trained on new data that excludes this digit. Based on prior work \citep{feigenbaum1961simulation, crowder2014principles}, we expect it may exhibit some decrease in performance as new training data interferes with its ability to correctly categorize the chosen digit; however, the decrease in performance will not be catastrophic.

\subsection{Results}

\begin{figure}[t!]
\centering
\includegraphics[width=0.8\textwidth]{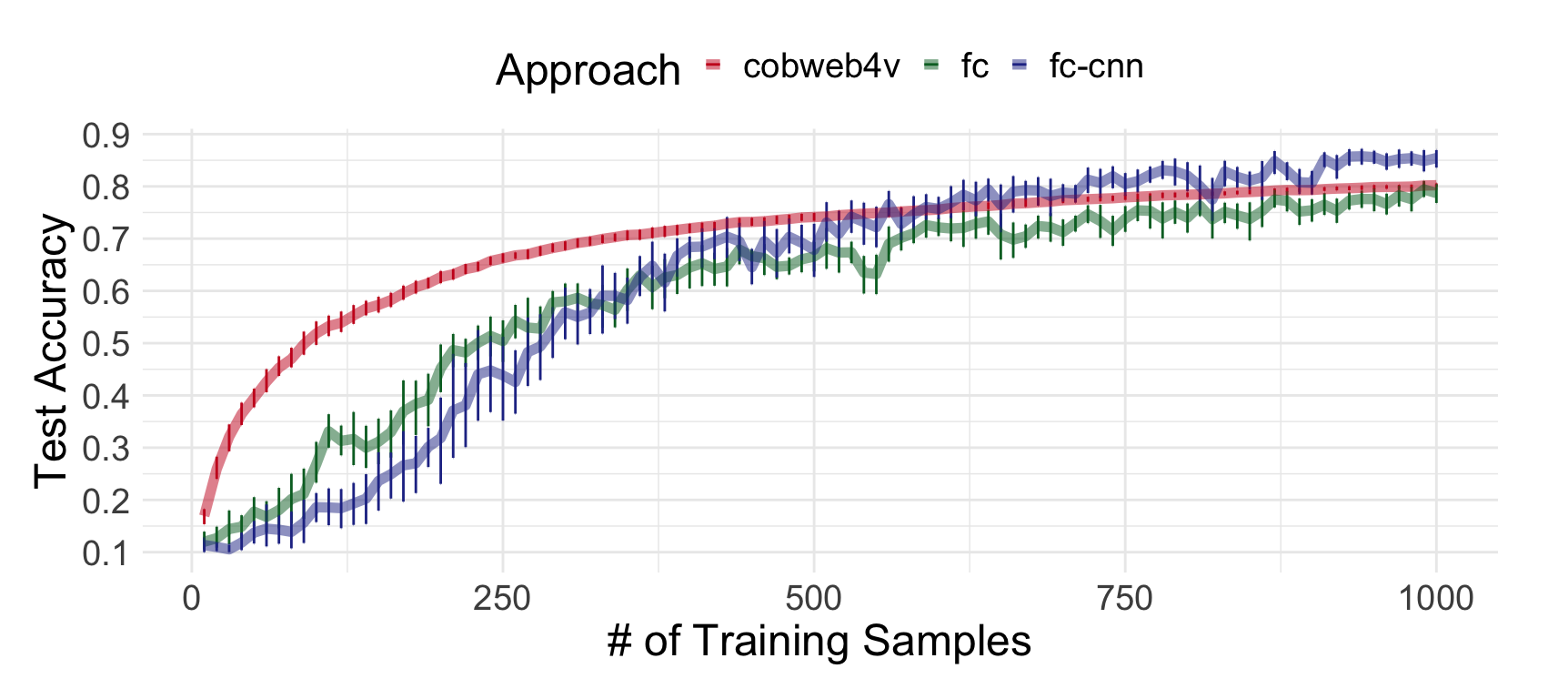}
\caption{Average test accuracy with 95\% confidence intervals on the MNIST test set as the number of training examples used increased in Experiment 1. We only present the learning curve of each approach after training with 1,000 examples.}
\label{fig:exp1}
\end{figure}

Figure~\ref{fig:exp1} shows the results of Experiment 1 for the first 100 splits (1,000 examples). We focus on this portion of the learning curve, as it highlights the difference between the approaches when only a small amount of data is available. Notably, \texttt{cobweb4v} demonstrates a significantly faster learning rate compared to the neural network baselines and reaches a commendable accuracy rate of 0.802 after 1,000 examples. The neural networks learn at a slower rate than \texttt{cobweb} but eventually achieve notable accuracy also (0.787 for \texttt{fc} and 0.854 for \texttt{fc-cnn} after 1,000 examples). At the end of training (after 60,000 examples), we found that over the 10 runs \texttt{cobweb4v} achieved a final mean test accuracy of $0.9514$ ($sd=0.00245$), \texttt{fc} achieved a mean accuracy of $0.9513$ ($sd=0.00941$), and \texttt{fc-cnn} achieved a mean accuracy of $0.9735$ ($sd=0.0072$). Compared to neural nets, \texttt{cobweb4v} reaches a commendable accuracy, comparable to the \texttt{fc} approach, while exhibiting much more stable behavior (far less variation across runs).

\begin{figure}[t!]
\centering
\includegraphics[width=0.8\textwidth]{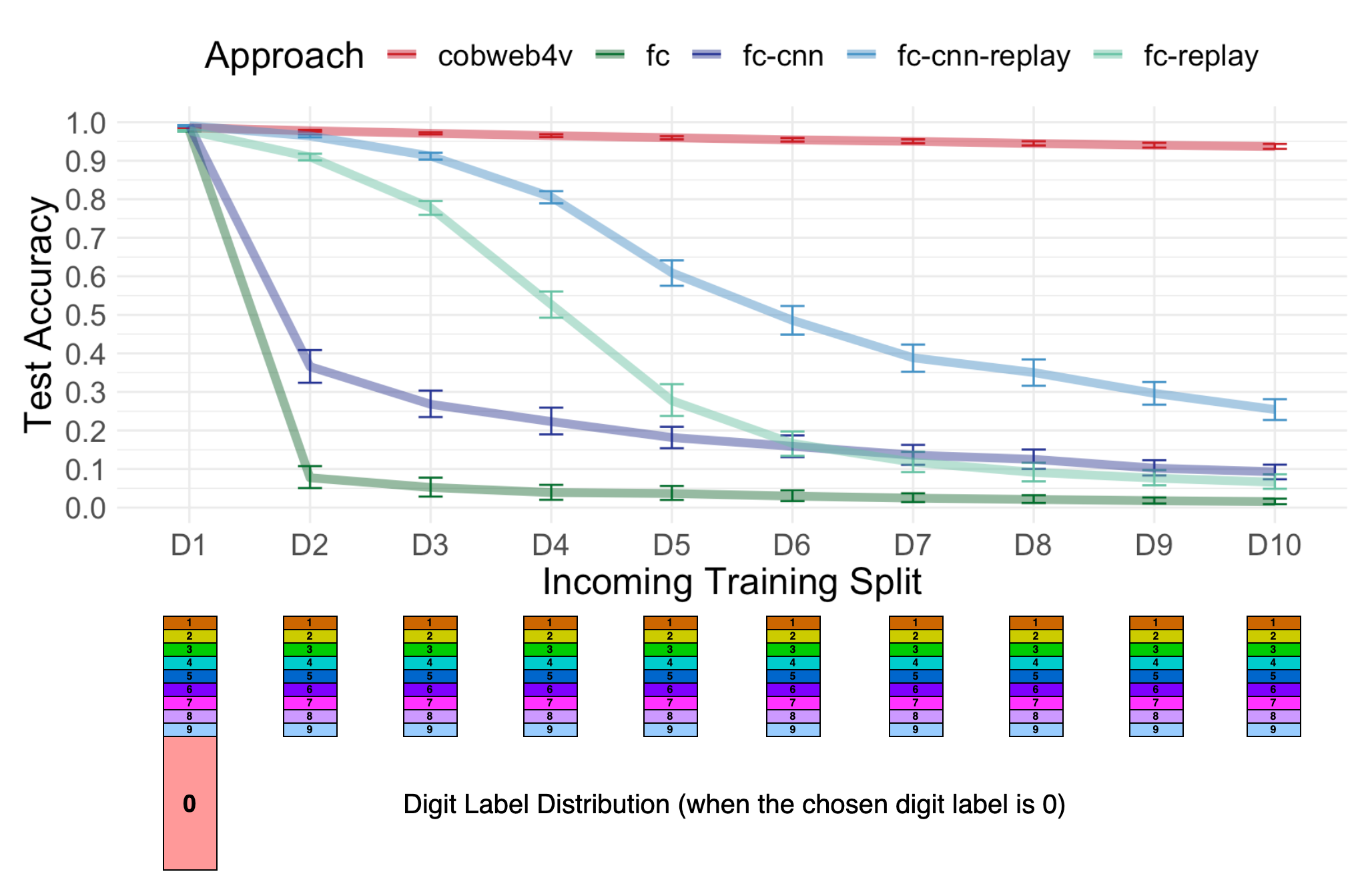}
\caption{Average test accuracy (with 95\% confidence intervals) on the chosen digit images from the MNIST test set after each training split (\texttt{D1}, ..., \texttt{D10}) in Experiment 2. In this experiment, the initial split \texttt{D1} includes all the training data from the chosen label, accompanied by 600 images from each non-chosen label. The subsequent 9 splits are generated by randomly and evenly dividing the remaining training data. The color blocks under the x-axis represent the digit distribution in each split when the chosen digit label is 0.}
\label{fig:exp2}
\end{figure}

Figure~\ref{fig:exp2} shows the results of Experiment 2, illustrating the variations in the summarized test accuracy of each approach on the chosen-digit test set after processing each of the 10 training splits. Despite all approaches learning all available digits after the first split, the accuracy of neural network baselines without replay still experiences a sharp decline and finally reaches an accuracy near zero. With replay, the neural networks have less performance decay initially, but as hypothesized, they also catastrophically forget when presented with an increasing number of splits. In contrast, the accuracy of \texttt{cobweb4v} gradually declines due to feature interference, where concepts at a branch point share similar features, which makes it difficult to determine which branch to follow during categorization. Unlike neural networks, which propagate activation over all nodes, Cobweb only activates a relatively small portion of its nodes during categorization ($N_{max}$). This is more computationally efficient, but may make it more susceptible to interference effects. The results show that \texttt{cobweb4v} maintains a respectable accuracy of 0.937 after training on 9 splits that do not contain the target digit, which suggests that, despite some interference, Cobweb is robust to catastrophic forgetting. Further, it outperforms all neural network approaches, even those using replay.

\subsection{Discussion}

In Experiment 1, we examine \texttt{cobweb4v}'s ability to learn visual concepts with varying numbers of training samples. The results show it is data efficient, learning more quickly than the neural net methods. We also find that it achieves comparable performance to the \texttt{fc} approach. The \texttt{fc-cnn} system ultimately achieves slightly better accuracy than both \texttt{cobweb4v} and \texttt{fc}, but this is to be expected as it has an additional representation learning approach (convolutional layers) that has been shown to increase performance on this task. This can be explained by the effectiveness of intermediate representations, which as mentioned, are more effective for generalization compared to the individual pixels.
We hypothesize that if Cobweb also utilized a representation learning technique, it would also achieve similar higher performance.
Some preliminary efforts have started to explore this research direction \citep{maclellan2022convolutional}. 
It is worth noting that, in this experiment, the neural networks utilize a smaller batch size of 5, which was necessary given the structure of the incremental learning task. While a larger batch size might enhance performance, it may compromise generalization ability, as highlighted by \cite{keskar2016large}. In real-world scenarios, managing the information scale in continuous learning of visual concepts poses challenges. Cobweb/4V showcases human-like learning by rapidly acquiring and refining its learned structure with limited data. This underscores the efficiency of Cobweb/4V in learning visual concepts, aligning closely with human learning behavior \citep{langley2022computational}. 

The results of Experiment 2 show that neural networks experience significant performance decay after learning the chosen digit due to catastrophic forgetting. Additionally, we observe that neural nets utilizing replay---a strategy anticipated to mitigate the forgetting effect---exhibit less initial forgetting, but eventually forget catastrophically as the number of splits increases. 
In contrast, \texttt{cobweb4v} avoids catastrophic forgetting, displaying only limited interference as it experiences more of the non-chosen digits. 
It is important to note that while Cobweb is instance-based, it also uses prototypes (the intermediate concepts) to generate predictions. In fact, when $N_{max}$ is small relative to  the size of the tree, then it may not use any leaf concepts (which represent specific instances) during prediction. Thus, there is no guarantee that Cobweb will be robust against forgetting. Cobweb uses intermediate concepts as an index to guide dynamic access to its tree structure during categorization. As mentioned previously, sometimes interference leads to poor categorization decisions that might prevent it from accessing the relevant portions of its tree. This provides a computational explanation for what forgetting is---a loss of access to previously available memories due to interference.  Therefore, it is plausible that Cobweb may still encounter challenges related to catastrophic forgetting.
Nevertheless, Cobweb/4V consistently outperforms all neural network baselines and retains concepts effectively, highlighting Cobweb/4V's continual learning superiority.

In light of our experimental findings, the unique attributes exhibited by Cobweb/4V in the context of continual learning deserve particular attention. 
On one hand, Cobweb/4V dynamically adapts its structure based on the incoming data to minimize the impact of forgetting, which echoes the principles of architectural-based methods reviewed in our study. For example, Cobweb/4V approach shares some similarity with neural network approaches like cascade correlation \citep{fahlman1989cascade}, hinting at their potential for mitigating catastrophic forgetting. On the other hand, Cobweb/4V shares some alignment with replay-based methods leveraging its tree structure to selectively revisit previously learned concepts that are relevant to the incoming instances. We believe Cobweb/4V's approach might suggest new strategies for replay, such as retaining prototypes or prototypical examples (akin to Cobweb's intermediate concept nodes) in the buffer. Our hope in the current study is that we might extract key lessons from Cobweb/4V's approach that might generalize to other machine learning methods, such as neural networks.

\section{Conclusion and Future Work}
In this study, we introduced Cobweb/4V, an novel extension of Cobweb that integrates computer vision concepts with human-like concept formation to support vision tasks. Our experiments assess Cobweb/4V's learning and performance suggesting it can both learn quickly and achieve good performance, indicating the potential efficacy of a modular, instance-based approach over a neural network representation for achieving more data-efficient learning. Additionally, our exploration of catastrophic forgetting in Cobweb/4V demonstrated its resistance to this phenomenon, displaying only slight performance decay due to interference, highlighting Cobweb/4V’s inherent advantages in the realm of continual learning, underscoring its stability and robustness.
In future work, we hope to explore the translation of other input processing ideas, such as convolutional and attentional processing, into Cobweb and to investigate if it achieves higher performance (similar to \texttt{fc-cnn} in Experiment 1) while staying robust to catastrophic forgetting during continual learning.
In conclusion, our research unveils promising avenues for the study of continual learning; we hope this will inspire further research into human-inspired machine learning approaches.


{\parindent -10pt\leftskip 10pt\noindent
\bibliographystyle{cogsysapa}
\bibliography{format}

}


\end{document}